\documentclass[a4paper,twoside]{article}
\usepackage{apalike}
\usepackage{times}  
\usepackage{helvet}  
\usepackage{courier}  
\usepackage[hyphens]{url}  
\usepackage{graphicx} 
\usepackage{natbib}  
\usepackage{caption} 
\usepackage{epsfig}
\usepackage{subcaption}
\usepackage{calc}
\usepackage{amssymb}
\usepackage{amstext}
\usepackage{amsmath}
\usepackage{amsthm}
\usepackage{multicol}
\usepackage{pslatex}

\def\std{\sigma_{\mathrm{std}}}
\usepackage{times}
\usepackage{latexsym}
\newtheorem{theorem}{Theorem}
\def\E{\mathrm{I\!E}}

\usepackage{newfloat}
\usepackage{listings}
\usepackage{microtype}
\usepackage{url}            
\usepackage{booktabs}       
\usepackage{amsfonts}       
\usepackage{nicefrac}       
\usepackage{microtype}      
\usepackage{xcolor}         
\usepackage{amsmath}
\usepackage{amssymb}
\usepackage[ruled,vlined]{algorithm2e}
\usepackage{bbm}
\usepackage{setspace}
\usepackage{graphicx}
\usepackage{graphics}
\usepackage{subcaption}
\usepackage{listings}
\usepackage{rotating}
\usepackage{multicol}
\usepackage{subfiles}
\usepackage{SCITEPRESS}     

\begin{document}

\title{iRNN: Integer-only Recurrent Neural Network}

\author{\authorname{Eyyüb Sari, Vanessa Courville, Vahid Partovi Nia\thanks{Corresponding author}}
\affiliation{Huawei Noah's Ark Lab, Montreal Research Centre, 7101 Park Avenue, H3N1X9 Quebec, Canada}
\email{\{vahid.partovinia\}@huawei.com}
}



\abstract{Recurrent neural networks (RNN) are  used in many real-world text and speech applications. They include complex modules such as recurrence, exponential-based activation, gate interaction, unfoldable normalization, bi-directional dependence, and attention. The interaction between these  elements prevents running them on integer-only operations without a significant performance drop. Deploying RNNs that include layer normalization and attention on integer-only arithmetic is still an open problem. We present a quantization-aware training method for obtaining a highly accurate integer-only recurrent neural network  (iRNN). Our approach supports layer normalization, attention, and an adaptive piecewise linear approximation of activations (PWL), to serve a wide range of RNNs on various applications. The proposed method is  proven to work on RNN-based language models and challenging automatic speech recognition, enabling AI applications on the edge. Our iRNN maintains similar performance as its  full-precision counterpart, their deployment on smartphones improves the runtime performance by  $2\times$, and reduces the model size by $4\times$.}

\onecolumn \maketitle \normalsize \setcounter{footnote}{0} \vfill

\section{Introduction}
RNN \citep{rumelhart1986rnn} architectures such as LSTM \cite{hochreiter1997lstm} or GRU \cite{cho2014gru} are the backbones of many downstream  applications.
RNNs now are part of large-scale systems such as neural machine translation   \cite{chen2018translstm,wang2019accelerating} and on-device systems such as Automatic Speech Recognition (ASR) \cite{he2019streaming}. RNNs are still highly used architectures in academia and industry, and their efficient inference requires  more elaborated studies.

In many edge devices, the number of computing cores is limited to a handful of computing units, in which parallel-friendly  transformer-based models lose their advantage.  There have been several studies in quantizing transformers to adapt them for edge devices but RNNs are largely ignored. 
Deploying RNN-based chatbot, conversational agent, and ASR on edge devices with limited memory and energy requires further computational improvements.  The  8-bit integer neural networks quantization \citep{jacob2017quantization} for convolutional architectures (CNNs) is shown to be an almost free lunch  to tackle the memory, energy, and latency costs, with a negligible accuracy drop \citep{krishnamoorthi2018quantizing}. 

\begin{figure*}
\centering
\includegraphics[scale=0.65]{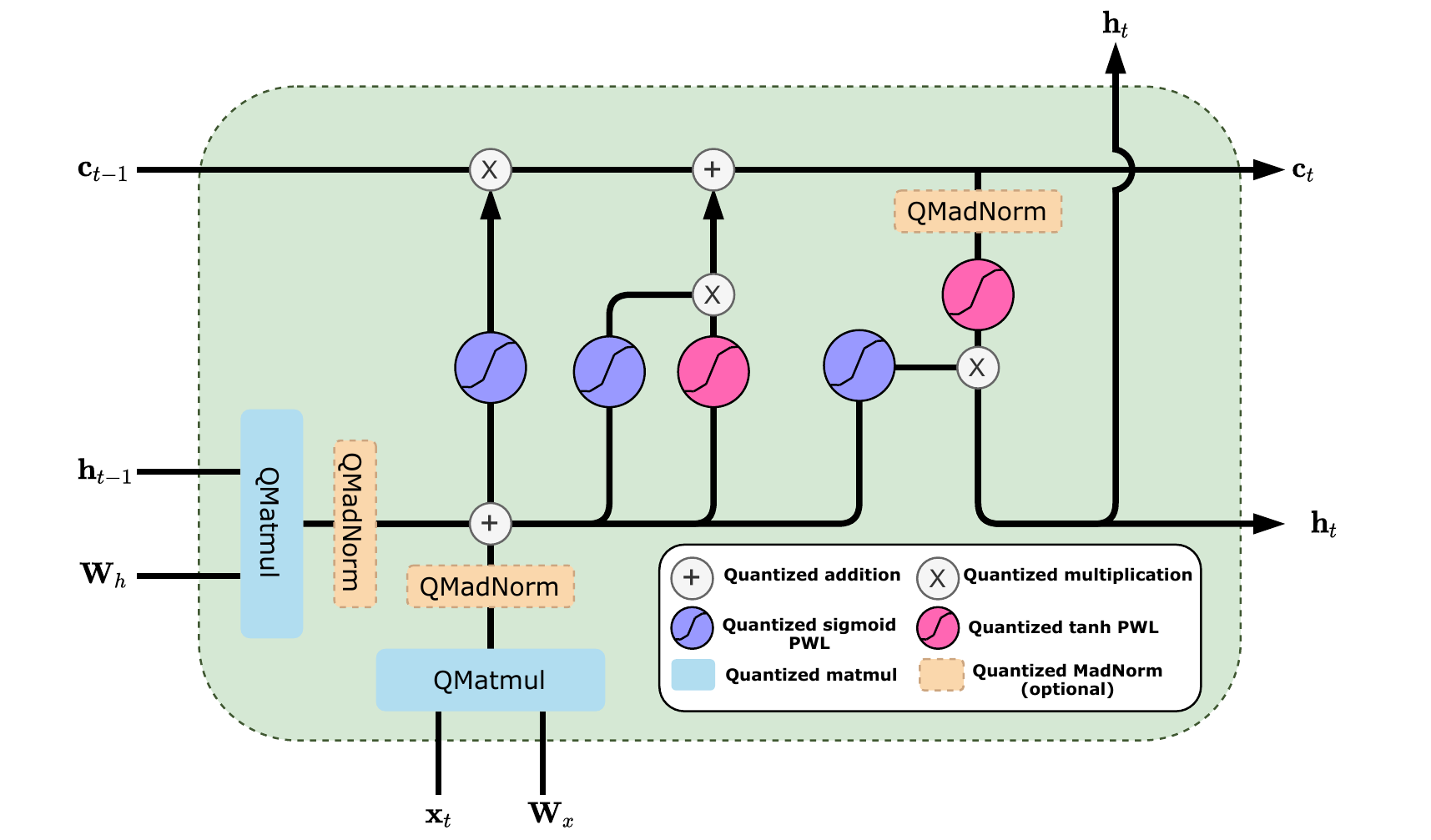}
\caption{Example of an integer-only LSTM cell (iLSTM). Layer normalization change to quantized integer friendly MadNorm (QMadNorm), full-precision matrix multiplications change to integer matrix multiplication (QMatmul), sigmoid and tanh activations are replaced with their corresponding piecewise linear (PWL) approximations. 
}
\label{fig:quantized-lstm}
\end{figure*}
Intuitively, quantizing RNNs is more challenging because the errors introduced by quantization will propagate in two directions, i) to the next layers, like feedforward networks ii) across timesteps. Furthermore, RNN cells are computationally more complex;  they include several element-wise additions and multiplications. They also have different activation functions that rely on the exponential function, such as sigmoid and hyperbolic tangent (tanh).

Accurate fully-integer RNNs calls for a new cell that is built using integer friendly operations. Our main motivation is to enable integer-only inference of RNNs on specialized edge AI computing hardware with no floating-point units, so we constrained the new LSTM cell to include only integer operations. 

First we build a fully integer  LSTM cell in which  its  inference require integer-only computation units, see Figure \ref{fig:quantized-lstm}. Our method can be applied to any RNN architecture, but here we focus on LSTM networks which are the most commonly used RNNs. 

Our contributions can be summarized as 
\begin{itemize}
\item providing a quantization-aware piecewise linear approximation algorithm to replace exponential-based activation functions (e.g. sigmoid and tanh) with integer-friendly  activation,
\item introducing an integer-friendly normalization layer based on mean absolute deviation,
\item proposing integer-only attention,
\item wrapping up these new modules into an LSTM cell towards  an integer-only LSTM cell.
\end{itemize}
We also implement our method on an anonymous smartphone, effectively showing $2\times$ speedup and $4\times$ memory compression. It is the proof that our method enables more RNN-based applications (e.g. ASR) on edge devices.

\section{Related Work}
With ever-expanding deep models, designing efficient neural networks enable wider adoption of deep learning in industry. Researchers recently started working on developing various quantization methods \citep{jacob2017quantization,courbariaux2016bnn,darabi2018bnnplus, esser2020lsq}. \citet{ott2016rnnlimitprec} explores low-bit quantization of weights for RNNs. They show binarizing weights lead to a massive accuracy drop, but ternarizing them keeps the model performance. \citet{courbariaux2016bnn} demonstrate quantizing RNNs to extremely low bits is challenging; they quantize weights and matrix product to 4-bit, but other operations such as element-wise pairwise and activations are computed in full-precision. 
\citet{hou2019normalization} quantize LSTM weights to 1-bit and 2-bit and show empirically that low-bit quantized LSTMs suffer from exploding gradients. Gradient explosion can be alleviated using normalization layers and leads to  successful training of low bit weights \cite{ardakani2018learning}.  \citet{sari2020normalization} studied the effect of normalization  in low bit networks theoretically,  and proved that low-bit training without  normalization operation is mathematically impossible; their work demonstrates the fundamental importance of involving normalization layers in quantized networks. 
 \citet{he2016effectivernn} introduce Bit-RNN and improve 1-bit and 2-bit RNNs quantization by constraining values within fixed range carefully; they keep activation computation and element-wise pairwise operations in full-precision. \citet{kapur2017lowprecrnn} build upon Bit-RNN and propose a low-bit RNN with minimal performance drop, but they increase the number of neurons to compensate for performance drop; they run activation and pair-wise operations on full-precision as well. 


\citet{wu2016gnmt} is a pioneering work in LSTM quantization, which demonstrated speed-up inference of large-scale LSTM models with limited performance drop by partially quantizing RNN cells. Their proposed method is tailored towards specific hardware. They use 8-bit integer for matrix multiplications and 16-bit integer for tanh, sigmoid, and element-wise operations but do no quantize attention  \citet{bluche2020sonosquantlstm}  propose an effective 8-bit integer-only LSTM cell for Keyword Spotting application on microcontrollers. They enforce weights and activations to be symmetric on fixed ranges $[-4, 4]$ and $[-1,1]$. This  prior assumption about the network's behaviour restrict generalizing their approach for wide range of RNN models. They propose a look-up table of 256 slots to represent the quantized tanh and sigmoid activations. However, the look-up table memory requirement explodes for bigger  bitwidth. Their solution does not serve complex tasks such as automatic speech recognition due to large look up table memory consumption. While demonstrating strong results on Keyword Spotting task, their assumptions on quantization range and bitwidth make their method task-specific. 
\section{Background}
We use the common linear algebra notation and use plain symbols to denote scalar values, e.g. $x\in \mathbb R$, bold lower-case letters to denote vectors, e.g. $\textbf{x} \in \mathbb{R}^{n}$, and bold upper-case letters to denote  matrices, e.g. $\textbf{X} \in \mathbb{R}^{m \times n}$. The element-wise multiplication is represented by $\odot$. 
\subsection{LSTM}
We define an LSTM cell as
\begin{align}
\begin{pmatrix} 
\mathbf{i}_t \\ 
\mathbf{f}_t \\ 
\mathbf{j}_t \\ 
\mathbf{o}_t 
\end{pmatrix} &= \mathbf{W}_x\mathbf{x}_t + \mathbf{W}_h\mathbf{h}_{t-1}, \label{eq:gates}\\
\mathbf{c}_t &= \sigma(\mathbf{f}_t) \odot \mathbf{c}_{t-1} + \sigma(\mathbf{i}_t) \odot \text{tanh}(\mathbf{j}_t), \label{eq:lstm-cell-state}\\
\mathbf{h}_t &= \sigma(\mathbf{o}_t) \odot \text{tanh}(\mathbf{c}_t),
\label{eq:vanilla-lstm}
\end{align}
where $\sigma(\cdot)$ is the sigmoid function; $n$ is the input hidden units dimension, and $m$ is the state hidden units dimension; $\mathbf{x}_t \in \mathbb{R}^{n}$ is the input for the current timestep $t \in \{1, ..., T\}$; $\mathbf{h}_{t-1} \in \mathbb{R}^{m}$ is the hidden state from the previous timestep and $\mathbf{h}_0$ is initialized with zeros; $\mathbf{W}_x \in \mathbb{R}^{4m \times n}$ is the input to state weight matrix; $\mathbf{W}_h \in \mathbb{R}^{4m \times m}$ is the  state to state weight matrix; $\{\mathbf{i}_t, \mathbf{f}_t, \mathbf{o}_t\} \in \mathbb{R}^{m}$ are the pre-activations to the \{input, forget, output\} gates; $\mathbf{j}_t \in \mathbb{R}^{m}$ is the pre-activation to the cell candidate; $\{\mathbf{c}_t, \mathbf{h}_t\} \in \mathbb{R}^{m}$ are the cell state and the hidden state for the current timestep, respectively. We omit the biases for the sake of notation simplicity. 
For a bidirectional LSTM (BiLSTM) the output hidden state at timestep $t$  is the concatenation of the forward  hidden state $\overrightarrow{\mathbf{h}}_t$ and the backward  hidden state $\overleftarrow{\mathbf{h}}_t$, $[\overrightarrow{\mathbf{h}}_t; \overleftarrow{\mathbf{h}}_t]$.

\subsection{LayerNorm}
\label{sec:layernorm}
Layer normalization \cite{ba2016layernorm} standardizes inputs across the hidden units dimension   with zero location and unit scale. Given hidden units $\mathbf{x} \in \mathbb{R}^H$, LayerNorm is defined as 


\begin{align}
\mu &= \frac{1}{H}\sum_{i=1}^H x_i, && \hat{x}_i = x_i - \mu \label{eq:centered-input} \\
\std^2 &= \frac{1}{H}\sum_{i=1}^H\hat{x}_i^2, && \std = \sqrt{\std^2} \label{eq:std}\\ 
\mathrm{LN}(\mathbf{x})_i &= y_i = \frac{\hat{x_i}}{\std} \label{eq:normalization}
\end{align}

where $\mu$ (\ref{eq:centered-input}) is the hidden unit mean, $\hat{x}_i$ (\ref{eq:centered-input}) is the centered hidden unit $x_i$, $\std^2$ (\ref{eq:std}) is the hidden unit variance, and $y_i$ (\ref{eq:normalization}) is the normalized hidden unit. In practice, one can scale $y_i$ by a learnable parameter $\gamma$ or shift by a learnable parameter $\beta$. The LayerNormLSTM cell is defined as  in \citet{ba2016layernorm}.


\subsection{Attention}
Attention is often used in encoder-decoder RNN architectures \citep{bahdanau2015attention, chorowski2015speechattention, wu2016gnmt}. We employ Bahdanau attention, also called additive attention \citep{bahdanau2015attention}. The attention mechanism allows the decoder network to attend to the variable-length output states from the encoder based on their relevance to the current decoder timestep. At each of its timesteps, the decoder extracts information from the encoder's states and summarizes it as a context vector, 
\begin{align}
\mathbf{s}_t &= \sum_{i=1}^{T_{\mathrm{enc}}} \alpha_{ti}\odot\mathbf{h}_{\mathrm{enc}_i} \\
\alpha_{ti} &= \frac{\text{exp}(e_{ti})}{\sum_{j=1}^{T_{\mathrm{enc}}}\text{exp}(e_{tj})} \label{eq:attn-alpha} \\
e_{ti} &= \mathbf{v}^\top \text{tanh}(\mathbf{W}_q\mathbf{h}_{t-1} + \mathbf{W}_k\mathbf{h}_{enc_i}) \label{eq:attn-qk}
\end{align}
where $\mathbf{s}_t$ is the context at decoder timestep $t$ which is a weighted sum of the encoder hidden states outputs $\mathbf{h}_{\mathrm{enc}_i} \in \mathbb{R}^{m_{\mathrm{enc}}}$ along encoder timesteps $i \in \{1, ..., T_{\mathrm{enc}}\}$ 
; $0<\alpha_{ti}<1$ are the attention weights attributed to each encoder hidden states based on the alignments $e_{ti} \in \mathbb{R}$; $m_{\mathrm{dec}}$ and $m_{\mathrm{enc}}$ are respectively the decoder and encoder hidden state dimension; 
$\{\mathbf{W}_q \in \mathbb{R}^{m_{\mathrm{att}} \times m_{\mathrm{dec}}}, \mathbf{W}_k \in \mathbb{R}^{m_{\mathrm{att}} \times m_{\mathrm{enc}}} \}$ are the weights matrices of output dimension $m_{\mathrm{att}}$ respectively applied to the query $\mathbf{h}_{t-1}$ and the keys $\mathbf{h}_{\mathrm{enc}_i}$; $\mathbf{v} \in \mathbb{R}^{m_{\mathrm{att}}}$ is a learned weight vector. The context vector is incorporated into the LSTM cell by modifying (\ref{eq:gates}) to 
\begin{eqnarray}
\begin{pmatrix} 
\mathbf{i} \\ 
\mathbf{f} \\ 
\mathbf{j} \\ 
\mathbf{o} 
\end{pmatrix} = \mathbf{W}_x\mathbf{x}_t + \mathbf{W}_h\mathbf{h}_{t-1} + \mathbf{W}_{s}\mathbf{s}_t
\end{eqnarray}
where $\mathbf{W}_{s} \in \mathbb{R}^{4m_{\mathrm{dec}} \times m_{\mathrm{enc}}}$. 
\subsection{Quantization}
\label{sec:background-quantization}
Quantization is a process whereby an input set is mapped to a lower resolution discrete set, called the quantization set $\mathcal{Q}$. The mapping is either performed from floating-points to integers (e.g. float32 to int8) or from a dense integer to another integer set with lower cardinality, e.g. int32 to int8. We follow the Quantization-Aware Training (QAT) scheme described in \citet{jacob2017quantization}. 

Given $x \in [x_{\min}, x_{\max}]$, we define the quantization process as
\begin{align}
q_x = \mathrm{q}(x) &= \Big\lfloor \frac{x}{S_x} \Big\rceil + Z_x \label{eq:quant}\\ 
r_x = \mathrm{r}(x) &= S_x(q_x - Z_x) \label{eq:dequant} \\
S_x = \frac{x_{\max} - x_{\min}}{2^b - 1},~~ & Z_x = \Big\lfloor \frac{ - x_{\min}}{S_x} \Big\rceil 
\end{align}
where the input is clipped between $x_{\min}$ and $x_{\max}$ beforehand; $\lfloor . \rceil$ is the round-to-nearest function; $S_x$ is the scaling factor (also known as the step-size); $b$ is the bitwidth, e.g. $b=8$ for 8-bit quantization, $b=16$ for 16-bit quantization; $Z_x$ is the zero-point corresponding to the quantized value of 0 (note that zero should always be included in $[x_{\min}, x_{\max}]$); $\mathrm{q}(x)$ quantize $x$ to an integer number and $\mathrm{r}(x)$ gives the floating-point value $\mathrm{q}(x)$ represents, i.e. $\mathrm{r}(x) \approx x$. We refer to $\{x_{\min}, x_{\max}, b, S_x, Z_x\}$ as \textbf{quantization parameters} of $x$.  Note that for inference, $S_x$ is expressed as a fixed-point integer number rather than a floating-point number, allowing for integer-only arithmetic computations \citep{jacob2017quantization}.


\section{Methodology}
In this section, we describe our task-agnostic quantization-aware training method to enable integer-only RNN (iRNN). 

\subsection{Integer-only activation}
\label{sec:quant-act}

First, we need to compute activation functions without relying on floating-point operations to take the early step towards an integer-only RNN. At inference, the non-linear activation is applied to the quantized input $q_x$, performs operations using integer-only arithmetic and outputs the quantized result $q_y$. Clearly, given the activation function $f$, $q_y = q(f(q_x))$;
as the input and the activation output are both quantized, we obtain a discrete mapping from $q_x$ to $q_y$. There are several ways to formalize this operation. The first  solution is a Look-Up Table (LUT), where $q_x$ is the index and $q_y = \text{LUT}[q_x]$. Thus, the number of slots in the LUT is $2^b$ (e.g. 256 bytes for $b=8$ bits input $q_x$). This method does not scale to large indexing bitwidths, e.g. 65536 slots need to be stored in memory for 16-bit activation quantization. LUT is not cache-friendly for large numbers of slots. The second solution is approximating the full-precision activation function using a fixed-point integer Taylor approximation, but the amount of computations grows as the approximation order grows. We propose to use a \textbf{Quantization-Aware PWL} that selects PWL knots during the training process to produce the linear pieces. 
Therefore the precision of approximation adapts to the required range of data flow automatically and provides highly accurate data-dependent activation approximation with fewer pieces.

\begin{figure}
\centering
\includegraphics[width=0.45\linewidth]{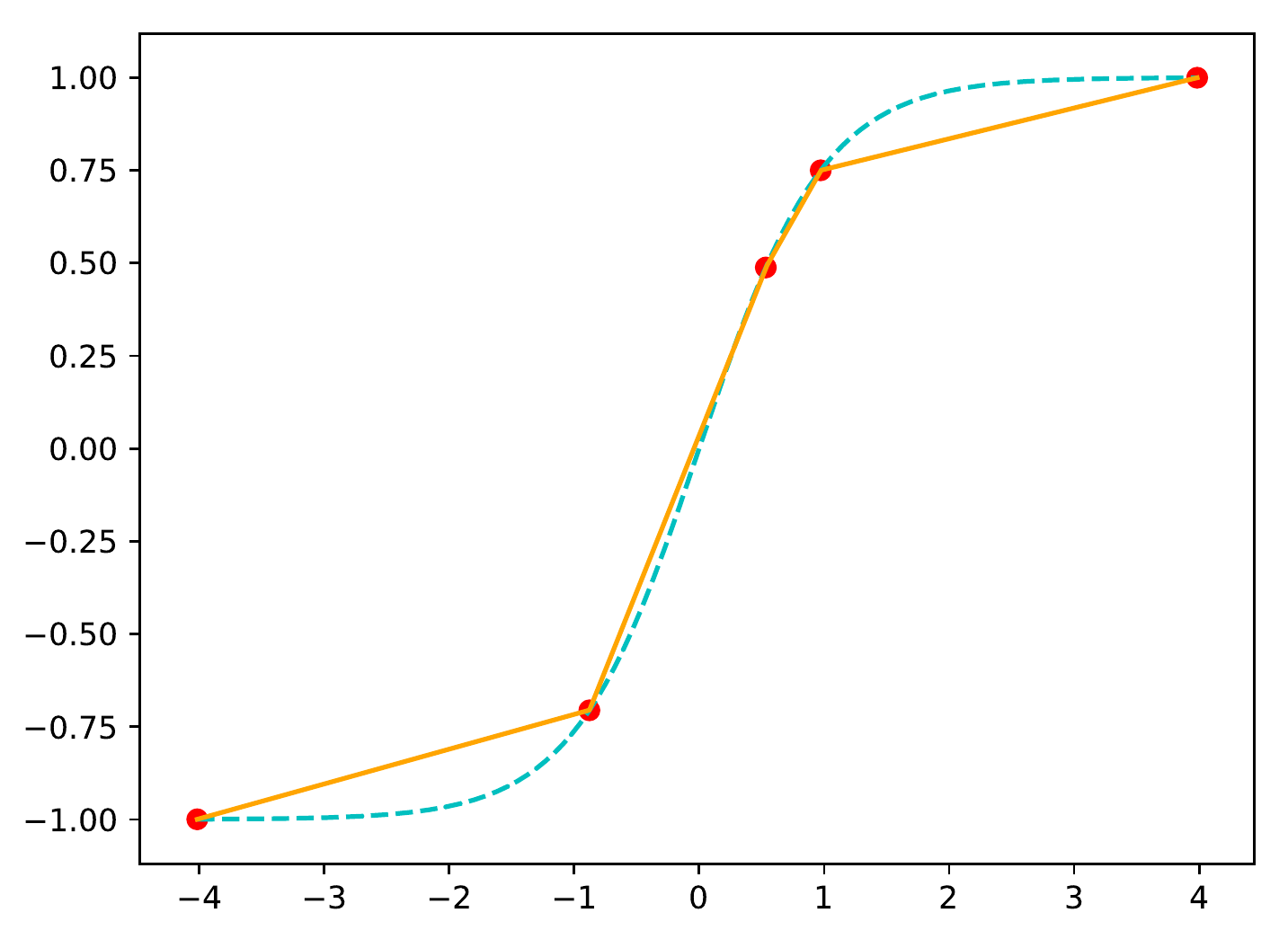}
\includegraphics[width=0.45\linewidth]{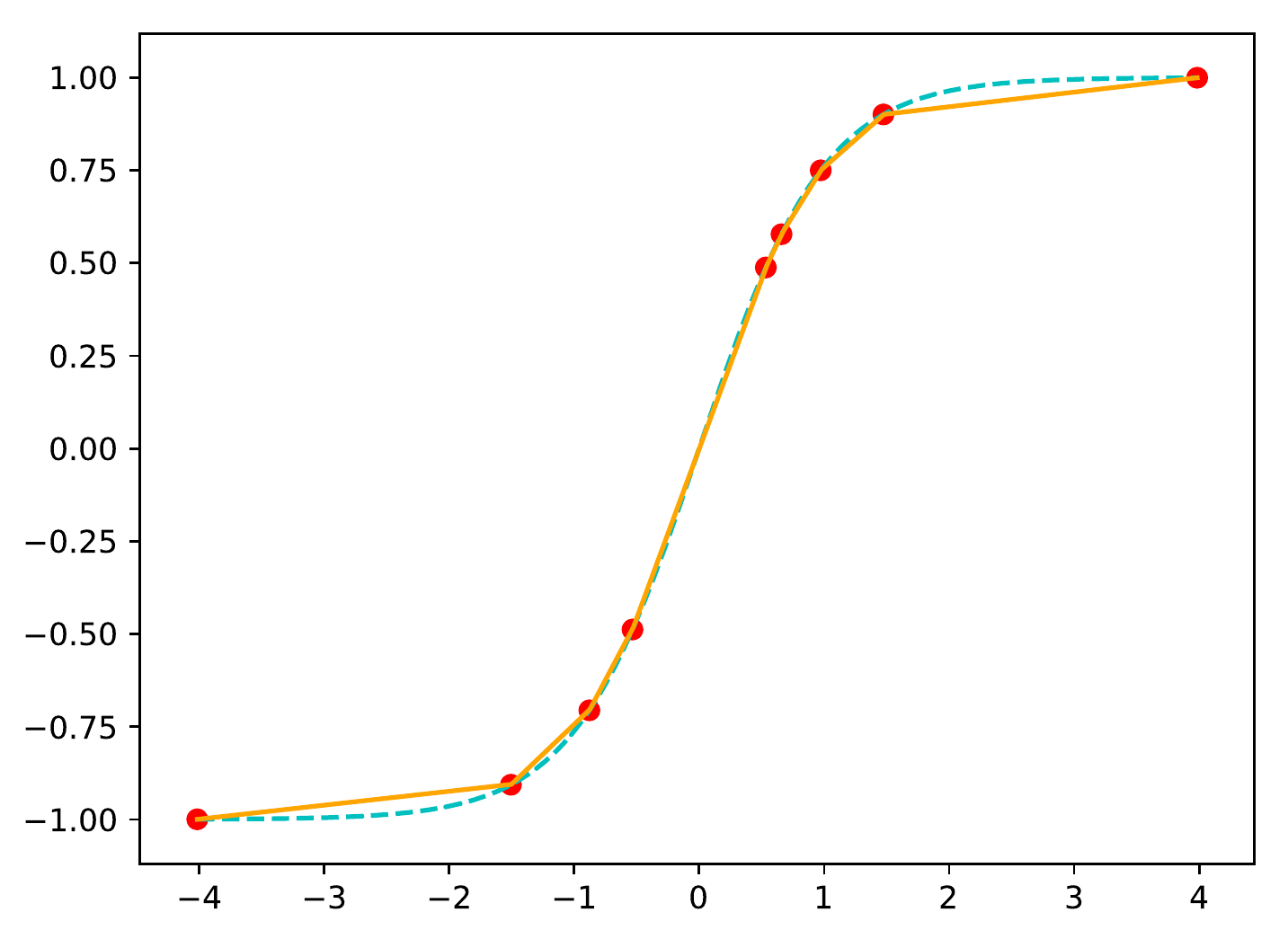}
\caption{
Tanh approximations using quantization-aware PWLs with 4 knots (left panel), 16 pieces (right panel) using (\ref{eq:pwl}). The dashed cyan curves are the true tanh functions, while the solid orange curves are its approximation from Algorithm \ref{alg:pwl}. Red dots are the knots. The more we add pieces, the better the approximation is. Our algorithm is able to prioritize sections of the function with more curvatures.
}
\label{fig:tanh-pwl}
\end{figure}

A PWL is defined as follows,
\begin{eqnarray}
g(x) = \sum_{i=1}^{N} \mathbbm{1}_{[k_i, k_{i+1})} \Big(a_i(x - k_i) + b_i \Big), \label{eq:pwl}
\end{eqnarray}
where $N$ is the number of linear pieces defined by $N + 1$ knots (also known as cutpoints or breakpoints); $\{a_i, k_i, b_i=f(k_i)\}$ are the slope, the knot, and the intercept of the $i^{\text{th}}$ piece respectively;  $\mathbbm{1}_{A}(x) = 1$ is the indicator function on $A$.  The more the linear pieces, the better the activation approximation is (see Figure \ref{fig:tanh-pwl}). A PWL is suitable for simple fixed-point integer operations. It only relies on basic arithmetic operations and is easy to parallelize because the computation of each piece is independent. Therefore, the challenge is to select the knot locations that provide the best PWL approximation to the original function $f$. Note in this regime, we only approximate the activation function on the subset corresponding quantized inputs and not the whole full-precision range. In our proposed method if $x=k_i$ then $g(x) = g(k_i) = b_i$, i.e. recovers the exact output $f(k_i)$. Hence, if the PWL has $2^b$ knots (i.e. $2^b - 1$ pieces), it is equivalent to a look-up table representing the quantized activation function. Thus, we constraint the knots to be a subset of the quantized inputs of the function we are approximating (i.e. $\{k_i\}_{i=1}^{N+1} \subseteq \mathcal{Q}$). 

We propose a recursive greedy algorithm to locate the knots during the quantization-aware PWL. The algorithm starts with $2^b - 1$ pieces and recursively removes one knot at a time until it reaches the specified number of pieces. The absolute differences between adjacents slopes are computed, and the shared knot from the pair of slopes that minimizes the absolute difference is removed; see Appendix Figure \ref{fig:pwl-algo}. The algorithm is simple to implement and applied only once at a given training step; see Appendix Algorithm \ref{alg:pwl}. This algorithm is linear in time and space complexity with respect to the number of starting pieces and is generic, allowing it to cover various nonlinear functions. Note that the PWL is specific to a given set of quantization parameters, i.e. the quantization parameters are kept frozen after its creation.

At inference, the quantization-aware PWL is computed as follows
\begin{eqnarray}
q_y = \Big\lfloor \sum_{i=1}^{N} \mathbbm{1}_{[q_{k_i}, q_{k_{i+1}})} \Big(\frac{S_x a_i}{S_y}(q_x - q_{k_i}) + \frac{b_i}{S_y}\Big) \Big\rceil  + Z_y,\nonumber
\end{eqnarray}
where the constants are expressed as fixed-point integers.

\subsection{Integer-only normalization}
\label{sec:madnorm}
Normalization greatly helps the convergence of quantized networks \citep{hou2019normalization, sari2020normalization}. There is a plurality of measures of location and scale to define normalization operation. The commonly used measure of dispersion is the standard deviation to define normalization, which is imprecise and costly to compute on integer-only hardware. However, the mean absolute deviation (MAD) is integer-friendly and defined as 
\begin{equation}
d = \frac{1}{H}\sum_{i=1}^{H} |x_i - \mu|  = \frac{1}{H}\sum_{i=1}^{H} |\hat{x}_i|.
\end{equation}
While the mean minimizes the standard deviation, the median minimizes MAD. We suggest measuring deviation with respect to mean for two reasons: i) the median is computationally more expensive ii) the absolute deviation from the mean is closer to the standard deviation. For Gaussian data, the MAD is $\approx 0.8 \std$ so that it might be exchanged with standard deviation. We propose to LayerNorm in LSTM with MAD instead of standard deviation and refer to it as MadNorm., where \eqref{eq:normalization} is replaced by
\begin{equation}
y_i = \frac{\hat{x}_i}{d}.
\label{eq:mad-normalization}
\end{equation}
MadNorm involves simpler operations, as there is no need to square and no need to  take the square root while taking absolute value instead of these two operations  is much cheaper. The values $\{\mu, \hat{x}_i, d, y_i\}$ are 8-bit quantized and computed as follows
\begin{align}
q_{\mu} &= \Big\lfloor \frac{S_x}{S_\mu N}\Big(\sum_{i=1}^Nq_{x_i} - NZ_x\Big) \Big\rceil + Z_\mu,
\label{eq:qmean}\\
q_{\hat{x}_i} &=\Big\lfloor \frac{S_x}{S_{\hat{x}}}(q_{x_i} - Z_x) - \frac{S_\mu}{S_{\hat{x}}}(q_{\mu} - Z_\mu) \Big\rceil + Z_{\hat{x}},
\label{eq:qcentered}\\
q_d &= \Big\lfloor \frac{S_{\hat{x}}}{S_{d}N}\sum_{i=1}^N |q_{\hat{x}_i} -  Z_{\hat{x}}|  \Big\rceil + Z_{d},
\label{eq:qmad}\\
q_{y_i} &= \Big\lfloor \frac{\frac{S_{\hat{x}}}{S_y S_d} (q_{\hat{x_i}} - Z_{\hat{x}})}{\mathrm{\max}(q_d, 1)} \Big\rceil + Z_y,
\label{eq:qmadnorm}
\end{align}
where all floating-point constants can be expressed as fixed-point integer numbers, allowing for integer-only arithmetic computations. Note that (\ref{eq:qmean}-\ref{eq:qmadnorm}) are only examples of ways to perform integer-only arithmetic for MadNorm, and may change depending on the software implementation and the target hardware.
We propose to quantize $\{\mathbf{v}, \mathbf{W}_q, \mathbf{W}_k\}$ to 8-bit. The vectors $\mathbf{h}_{t-1}$ and $\mathbf{h}_{\mathrm{enc}_i}$ are quantized thank to the previous timestep and/or layer. The matrix multiplications in (\ref{eq:attn-qk}) are performed in 8-bit and their results are quantized to 8-bit, each with their own quantization parameters. Since those matrix multiplications do not share the same quantization parameters, the sum  (\ref{eq:attn-qk}) require proper rescaling and the result is quantized to 16-bit. We  found 8-bit quantization adds too much noise, thus preventing the encoder-decoder model to work correctly. The tanh function  in (\ref{eq:attn-qk}) is computed using quantization-aware PWL and its outputs are quantized to 8-bit.  The alignments $e_{ti}$ are quantized to 16-bit (\ref{eq:attn-qk}). The exponential function in $\alpha_{ti}$ is computed using a quantization-aware PWL, with its outputs quantized to 8-bit. We found that quantizing the softmax denominator (\ref{eq:attn-alpha}) to 8-bit introduces too much noise and destroys attention. 8-bit attention does not offer enough flexibility and prevents fine grained decoder attention to the encoder. We left the denominator in 32-bit integer value and defer quantization to 8-bit in the division. The context vector $\mathbf{s}_t$ is quantized to 8-bit. Note that in practice we shift the inputs to softmax for numerical stability reasons (i.e. $e_{ti} - \max\limits_j e_{tj}$).

\begin{theorem}[Scale convergence]
\label{theo:scale}
Suppose $X_i$ pairwise independent  samples from the same probability space $(\Omega, \mathcal F, \Pr)$ with $\mu=\E(X_i)$ and are absolutely integrable,  then $D_n={1\over n}\sum_{i=1}^n |X_i-\mu|$ converges almost surely to $\tilde \sigma = \E(|X-\mu|).$\\
\end{theorem}
{\bf Proof:} Absolutely integrable condition assures the existence of $\mu=\E(X_i)<\E(|X_i|) <\infty$ and hence the existence of $\E|X_i-\mu|<\E|X_i|+\mu <\infty$. The proof is straightforward by applying the standard strong law of large numbers to $Y_i= |X_i-\mu|. \blacksquare$

One may prove the central limit theorem by replacing absolute integrability with square integrability, and exchanging pairwise independence with mutual independence. 
Convergence to the population scale $\tilde \sigma$ in Theorem~\ref{theo:scale} paves the way to show that our MadNorm enjoys a concentration inequality similar to LayerNorm.
\begin{theorem}[Concentration inequality]
\label{theo:con}
Suppose the random variable $X$ with mean $\mu$ is absolutely integrable with respect to the probability measure $P$. Then for a positive $k$, 
A vanilla LSTM cell comprises  matrix multiplications, element-wise additions, element-wise multiplications, tanh, and sigmoid activations (\ref{eq:gates} - \ref{eq:vanilla-lstm}). We quantize the weights matrices $\mathbf{W}_x$ and $\mathbf{W}_h$ to 8-bit. The inputs $\mathbf{x}_t$ and hidden states $\mathbf{h}_{t-1}$ are already 8-bit quantized from the previous layer and from the previous timestep. The cell states $\mathbf{c}_t$ are theoretically unbounded (\ref{eq:lstm-cell-state}); therefore the amount of quantization noise potentially destroys the information carried by $\mathbf{c}_t$, if it spans a large range. When performing QAT on some pre-trained models, it is advised to quantize $\mathbf{c}_t$ to 16-bit. Therefore, $\mathbf{c}_t$  is 8-bit quantized unless stated otherwise but can be quantized to 16-bit if necessary. Matrix multiplications in (\ref{eq:gates}) are performed with 8-bit arithmetic, and their outputs are quantized to 8-bit based on their respective quantization parameters. The sum between the two matrix multiplications outputs in (\ref{eq:gates}) requires proper rescaling, because they do not share the same quantization parameters.
$$\Pr\left(\lvert {X-\mu \over \tilde \sigma} \rvert<k\right)\geq 1-{1\over k}$$
\end{theorem}
{\bf Proof:} 
Take $Y=\lvert {X-\mu \over \tilde \sigma} \rvert$. The random variable $X$ is absolutely integrable, and so is $Y$. 
\begin{eqnarray*}
\E(Y) &=& \int_0^\infty Y dP = \int_0^k YdP + \int_k^\infty YdP\\
&\geq& 0+ k\int_k^\infty YdP = k\Pr(Y>k), 
\end{eqnarray*}
it follows immediately that $\Pr\left( \lvert {X-\mu \over \tilde \sigma} \rvert>k\right)\leq {1\over k}. \blacksquare$\\
Theorem~\ref{theo:con} assures that independent of distribution of data, this MadNorm brings the mass of the distribution around the origin. This is somehow expected from any normalization method. It is not surprising to see that LayerNorm also has a similar property, and therefore in this sense LayerNorm and MadNorm assures the tail  probability far from the origin is negligible. 

There is a slight difference between the concentration inequality of LayerNorm and MadNorm. The LayerNorm provides a tighter bound, i.e. the bound in Theorem~\ref{theo:con} changes from $1-{1\over k}$ to $1-{1\over k^2}$ but it also requires more assumptions like the square integrability of $X$. 
\subsection{Integer-only attention}
\label{sec:qattention}
Attention plays a crucial role in modern encoder-decoder architectures. The decoder relies on attention to extract information from the encoder and provide predictions. Attention is the bridge between the encoder and the decoder. Careless quantization of attention breaks apart the decoder due to quantization noise. 

\subsection{Integer-only LSTM network}
\label{sec:qlstm}
The results of the sum are quantized to 8-bit; however, 16-bit quantization might be necessary for complex tasks.  The sigmoid and tanh activations in  (\ref{eq:lstm-cell-state}) and (\ref{eq:vanilla-lstm}) are replaced with their own quantization-aware PWL, and their output is always quantized to 8-bit. The element-wise multiplications operations are distributive, and sharing quantization parameters is not required. In (\ref{eq:lstm-cell-state}), the element-wise multiplications are quantized to 8-bit, but can be quantized to 16-bit if $\mathbf{c}_t$ is quantized to 16-bit as well; the element-wise additions are quantized based on $\mathbf{c}_t$'s bitwidth (i.e. 8-bit or 16-bit).

The element-wise multiplications between sigmoid and tanh in (\ref{eq:vanilla-lstm}) is always quantized to 8-bit, because $\mathbf{h}_t$ are always quantized to 8-bit. Following this recipe, we obtain an integer-only arithmetic LSTM cell, see Figure. \ref{fig:quantized-lstm}. For LSTM cells with LayerNorm quantized MadNorm layers are used instead of LayerNorm. Appendix \ref{sec:appendix-lstm-continue} 
provides details about quantization of other types of layers in an LSTM model.


%


\section{Experiments}
\label{sec:experiments}
We evaluate our proposed method, iRNN, on language modeling and automatic speech recognition. We also implemented our approach on a smartphone to benchmark inference speedup, see \ref{sec:inference}.
\subsection{Language modeling on PTB}
\begin{table}
{\small
\begin{center}
\begin{tabular}{ l | c | c }
      \hline 
	  LayerNorm LSTM  & val & test\\
	  \hline
	  Full-precision & $98.58 \pm 0.35$ & $ 94.84 \pm 0.21$ \\
	  \hline
	  \hline
      
      PWL4 & $101.40 \pm 0.70$ & $98.11 \pm 0.75$ \\ 
      PWL8 & $98.14 \pm 0.11$ & $95.03 \pm 0.16$ \\
      PWL16 & $98.09 \pm 0.06$ & $94.92 \pm 0.05$ \\ 
      PWL32 & $\mathbf{97.97} \pm 0.01$ & $\mathbf{94.81} \pm 0.02$ \\ 
\end{tabular} 
\end{center}
}
\caption{Word-level perplexities on PTB with a LayerNorm LSTM and quantized models with a different number of PWL pieces. LayerNorm is replaced with MadNorm for the quantized models (iRNN). Best results are averaged across 3 runs $\pm$ standard deviation.}
\label{table:ptb}
\end{table}
As a proof of concept, we perform several experiments on full-precision and  fully 8-bit quantized models on the Penn TreeBank (PTB) dataset \citep{marcus1993ptb}. We report perplexity per word as a performance metric.

For the quantized models, the LayerNorm is replaced with MadNorm. We do not train full-precision models with MadNorm to make our method comparable with common full-precision architectures. 
We can draw two conclusions from the results presented in Table \ref{table:ptb}, i) replacing LayerNorm by MadNorm does not destroy model performance, ii) using eight linear pieces is enough to retain the performance of the model, but adding more linear pieces improves the performance. We could obtain even superior results in the quantized model compared to the full-precision model because of the regularization introduced by quantization errors. 

\subsection{Language modeling on WikiText2}
\label{sec:wikitext2}
\begin{table}
{\small
\begin{center}
\begin{tabular}{ l | c | c }
    \hline
	  Mogrifier LSTM & val & test\\
	  \hline
	  Full-precision & $60.27 \pm 0.34$ & $58.02 \pm 0.34$ \\
	  \hline
	  \hline
       PWL8   & $ 60.91 \pm 0.04$ & $ 58.54\pm $ 0.07 \\
       PWL16 & $ 60.65 \pm 0.09 $ & $ 58.21 \pm  0.08$\\ 
      PWL32 & $ \mathbf{60.37} \pm 0.03$ & $ \mathbf{57.93} \pm 0.07$

\end{tabular} 
\end{center}
\caption{Word-level perplexities on WikiText2 with Mogrifier LSTM and quantized models with different number of PWL pieces. Best results are averaged across 3 runs $\pm$ standard deviations.}
\label{table:mogrifier-lstm}
}
 \end{table}

We evaluated our proposed method on the WikiText2 dataset \citep{merity2016wikitext} with a state-of-the-art RNN, Mogrifier LSTM \citep{melis2020mogrifier}. The original code\footnote{\url{https://github.com/deepmind/lamb}} was written in TensorFlow, we reimplemented our own version in PyTorch by staying as close as possible to the TensorFlow version. We follow the experimental setup from the authors\footnote{\url{https://github.com/deepmind/lamb/blob/254a0b0e330c44e00cf535f98e9538d6e735750b/lamb/experiment/mogrifier/config/c51c838b33a5+_tune_wikitext-2_35m_lstm_mos2_fm_d2_arms/trial_747/config}} as we found it critical to get similar results.
We use a two layer Mogrifier LSTM. The setup and hyper-parameters for the experiments can be found in Appendix \ref{sec:appendix-mogrifier} 
to save some space.    We present our results averaged over 3 runs in Table \ref{table:mogrifier-lstm}. We use the best full-precision model, which scores $59.95$ perplexity to initialize the quantized models. Our method is able to produce 8-bit quantized integer-only Mogrifier LSTM with similar performance to the full-precision model with only about $0.3$ perplexity increase for the quantized model with a PWL of 32 pieces and a maximum of about $0.9$ perplexity increase with a number of pieces as low as 8. Interestingly, a pattern emerged by doubling the number of pieces, as we get a decrease in perplexity by about $0.3$. We also perform a thorough ablation study of our method in Appendix Table \ref{table:mogrifier-ablation}. 
Surprisingly, we found that stochastic weight averaging for quantized models exhibits the same behavior as for full-precision models and improved performance thanks to regularization. While experiments on the PTB dataset were a demonstration of the potential of our method, these experiments on WikiText2 show that our proposed method is able to stay on par with state-of-the-art RNN models.

\subsection{ASR on LibriSpeech}
\begin{table}
{\small
\begin{center}
\begin{tabular}{ l | c | c | c}
    \hline
	  ESPRESSO LSTM & set & clean & other\\
	  \hline
	  Full-precision & dev & 2.99 & 8.77 \\
	  iRNN PWL96* & dev & 3.73 & 10.02 \\
	  \hline 
	  \hline 
	  Full-precision & test & 3.37 & 9.49 \\
      
      iRNN PWL96* & test & 4.11 & 10.71 \\

\end{tabular} 
\end{center}
\caption{WER\% on LibriSpeech with ESPRESSO LSTM (Encoder-Decoder LSTM with Attention) with LM shallow fusion. *(160 pieces were used for the exponential function)}
\label{tab:asr}
}\end{table}
ASR is a critical edge AI application, but also challenging due to the nature of the task. Voice is diverse in nature as human voice  may vary in pitch, accent, pronunciation style, voice volume, etc.  
While we showed our method is working for competitive language modeling task, one can argue ASR is a more practical and at the same time more difficult task for edge and IoT applications. Thefore, we experiment on an ASR task based on the setup of \citet{wang2019espresso} and their ESPRESSO framework\footnote{\url{https://github.com/freewym/espresso}}. We used an LSTM-based Attention Encoder-Decoder (ESPRESSO LSTM) trained on the strong ASR LibriSpeech dataset \citep{panayotov2015librispeech}. Experiments setup and hyper-parameters are provided in Appendix \ref{sec:appendix-asr}
We initialize the quantized model from the pre-trained full-precision ESPRESSO LSTM. In our early experiments, we found that quantizing the model to 8-bit would not give comparable results. After investigation, we noticed it was mainly due to two reasons, i) the cell states $\mathbf{c}_t$ had large ranges (e.g. $[-17, 15]$), ii) the attention mechanism was not letting the decoder attend the encoder outputs accurately. Therefore, we quantize the pre-activation gates (\ref{eq:gates}), the element-wise multiplications in (\ref{eq:lstm-cell-state}) and cell states $\mathbf{c}_t$ to 16-bit. The attention is quantized following our described integer-only attention method. Everything else is quantized to 8-bit following our described method. The quantized model has a similar performance to the full-precision model, with a maximum of $1.25$ WER\% drop (Table \ref{tab:asr}). We believe allowing the model to train longer would reduce the gap.

\subsection{Inference measurements}
\label{sec:inference}
\begin{table}
{
\begin{center}
\begin{tabular}{ l | c | c | c}
    \hline
	  LSTM &  ms & iter/s & speedup\\
	  \hline
	  Full-precision  & 130 & 7.6 & $1.00\times$ \\
	  \hline 
	  \hline
	  iRNN PWL32  & 84 & 11.8 & $1.54\times$ \\
	  iRNN PWL8   & 61 & 14.9 & $\mathbf{1.95\times}$ \\
	  iRNN without QAct  & 127 & 7.8 & $1.02\times$\\
\end{tabular} 
\end{center}
\caption{Inference measurements on an anonymous smartphone based on a custom fork from PyTorch 1.7.1. The model is a one LSTM cell with a state size of 400.}
\label{tab:inf}
}
\end{table}
We implemented an 8-bit quantized integer-only LSTM with PWL model based on a custom PyTorch \citep{paszke2019pytorch} fork from 1.7.1. We implemented an integer-only PWL kernel using NEON intrinsics. We benchmark the models on an anonymous smartphone using the $\mathrm{speed\_benchmark\_torch}$ tool\footnote{\url{https://github.com/pytorch/pytorch/blob/1.7/binaries/speed_benchmark_torch.cc}}. We warm up each model for 5 runs and then measure the inference time a hundred times and report an average. The sequence length used is 128, and the batch size is one. We benchmark our iRNN LSTM model using PWLs with 32 pieces, and 8 pieces which achieve up to $2\times$ speedup. We also evaluate our iRNN with full-precision computations (iRNN w/o QAct) for the activation where no speedup was observed for this state size. We believe it is due to round-trip conversions between floating-points and integers (Table \ref{tab:inf}). There is a lot of room for improvements to achieve even greater speedup, such as writing a \texttt{C++} integer-only LSTM cell, fusing operations, and better PWL kernel implementation.

\section{Conclusion}

We propose a task-agnostic and flexible methodology to enable integer-only RNNs. To the best of our knowledge, we are the first to offer an approach to quantize all existing operations in modern RNNs, supporting normalization and attention. We evaluated our approach on high-performance LSTM-based models on language modeling and ASR, which have distinct architectures and variable computation requirements. We show that RNN can be fully quantized while achieving similar performance as their full-precision counterpart. We benchmark our method on an anonymous smartphone, where we obtain $2\times$ inference speedup and $4\times$ memory reduction. This allows to deploy a wide range of RNN-based applications on edge and on specialized AI hardware and microcontrollers that lack floating point operation.





\bibliographystyle{apalike}
\bibliography{custom}

\section{Appendix}
\label{sec:appendix}

\subsection{Specific details on LSTM-based models}
\label{sec:appendix-lstm-continue}
For BiLSTM cells, nothing stated in section Integer-only LSTM network is changed except that we enforce the forward LSTM hidden state $\overrightarrow{\mathbf{h}}_t$ and the backward LSTM hidden state $\overleftarrow{\mathbf{h}}_t$ to share the same quantization parameters so that they can be concatenated as a vector. If the model has embedding layers, they are quantized to 8-bit as we found they were not sensitive to quantization. If the model has residual connections (e.g. between LSTM cells), they are quantized to 8-bit integers. In encoder-decoder models the attention layers would be quantized following section Integer-only attention. The model's last fully-connected layer's weights are 8-bit quantized to allow for 8-bit matrix multiplication. However, we do not quantize the outputs and let them remain 32-bit integers as often this is where it is considered that the model has done its job and that some postprocessing is performed (e.g. beam search). 

\subsection{Experimental details}
\begin{table*}
\begin{center}
\begin{tabular}{ l | c | c }
\hline
	  iRNN Mogrifier LSTM & val & test\\
	  \hline
	  
	  w/o PWL & $60.40 \pm 0.05$ & $ 57.90 \pm 0.01$  \\
    ~~w/o Quantized Activations & $60.40 \pm 0.03$ & $57.95 \pm 0.003$ \\ 
      ~~~w/o Quantized Element-wise ops& $60.08 \pm 0.10$ & $57.61 \pm 0.23$ \\
      ~~~~w/o Quantized Matmul & $60.10 \pm 0.05$ & $57.64 \pm 0.10$ \\ 
      ~~~~~w/o Quantized Weights (Full-precision) & $60.27 \pm 0.34$ & $58.02 \pm 0.34$ \\
\end{tabular} 
\end{center}
\caption{Ablation study on quantized Mogrifier LSTM training on WikiText2. iRNN w/o PWL is the quantized model using LUT instead of PWL to compute the activation function. Best results are averaged across 3 runs, and standard deviations are reported.}
\label{table:mogrifier-ablation}
\end{table*}

We provide a detailed explanation of our experimental setups.  
\subsubsection{LayerNorm LSTM on PTB}
\label{sec:appendix-ptb}
We provide detailed information about how the language modeling on PTB experiments are performed. The vocabulary size is 10k, and we follow dataset preprocessing as done in \citet{mikolov2012subword}. We report the best \emph{perplexity} per word on the validation set and test set for a language model of embedding size 200 with one LayerNormLSTM cell of state size 200. The lower the perplexity, the better the model performs. In these experiments, we are focusing on the relative increase of perplexity between the full-precision models and their 8-bit quantized counterpart. We did not aim to reproduce state-of-the-art performance on PTB and went with a naive set of hyper-parameters. The full-precision network is trained on for 100 epochs with batch size 20 and BPTT \citep{werbos1990bptt} window size of 35. We used the SGD optimizer with weight decay of $10^{-5}$ and learning rate 20, which is divided by 4 when the loss plateaus for more than 2 epochs without a relative decrease of $10^{-4}$ in perplexity. We use gradient clipping of 0.25. We initialize the quantized models from the best full-precision checkpoint and train from another 100 epochs. For the first 5 epochs we do not enable quantization to gather range statistics to compute the quantization parameters.
\subsubsection{Mogrifier LSTM on WikiText2}
\label{sec:appendix-mogrifier}
We describe the experimental setup for Mogrifier LSTM on WikiText2. Note that we follow the setup of \citet{melis2020mogrifier} where they do not use dynamic evaluation \citep{krause2018dyneval} nor Monte Carlo dropout \citep{gal2016mcdropout}.
The vocabulary size is 33279. We use a 2 layer Mogrifier LSTM with embedding dimension 272, state dimension 1366, and capped input gates. We use 6 modulation rounds per Mogrifier layer with low-rank dimension 48. We use 2 Mixture-of-Softmax layers \citep{yang2018mos}. The input and output embedding are tied.  We use a batch size of 64 and a BPTT window size of 70. We train the full-precision Mogrifier LSTM for 340 epochs, after which we enable Stochastic Weight Averaging (SWA) \citep{izmailov2018swa} for 70 epochs. For the optimizer we used Adam \citep{kingma2014adam} with a learning rate of $\approx 3\times 10^{-3}$, $\beta_1=0$, $\beta_2=0.999$ and weight decay $\approx 1.8\times 10^{-4}$. We clip gradients' norm to 10. We use the same hyper-parameters for the quantized models from which we initialize with a pre-trained full-precision and continue to train for 200 epochs. During the first 2 epochs, we do not perform QAT, but we gather min and max statistics in the network to have a correct starting estimate of the quantization parameters. After that, we enable 8-bit QAT on every component of the Mogrifier LSTM: weights, matrix multiplications, element-wise operations, activations. Then we replace activation functions in the model with quantization-aware PWLs and continue training for 100 epochs.  

We perform thorough ablation on our method to study the effect of each component.  Quantizing the weights or the weights and matrix multiplications covers about $0.1$ of the perplexity increase. There is a clear performance drop after adding quantization of element-wise operations with an increase in perplexity of about $0.3$. This is both due to the presence of element-wise operations in the cell and hidden states computations affecting the flow of information across timesteps and to the residual connections across layers. On top of that, adding quantization of the activation does not impact the performance of the network.
 \subsubsection{ESPRESSO LSTM on LibriSpeech}
   
	  

 \label{sec:appendix-asr}
 The encoder is composed of 4 CNN-BatchNorm-ReLU blocks followed by 4 BiLSTM layers with 1024 units. The decoder consists of 3 LSTM layers of units 1024 with Bahdanau attention on the encoder's hidden states and residual connections between each layer. The dataset preprocessing is exactly the same as in \citet{wang2019espresso}. 
 We train the model for 30 epochs on one V100 GPU, which takes approximately 6 days to complete. We use a batch size of 24 while limiting the maximum number of tokens in a mini-batch to 26000. Adam is used with a starting learning rate of $0.001$, which is divided by 2 when the validation set metric plateaus without a relative decrease of $10^{-4}$ in performance. Cross-entropy with uniform label smoothing  $\alpha=0.1$ \citep{szegedy2016inceptionv3} is used as a loss function. At evaluation time, the model predictions are weighted using a pre-trained full-precision 4-layer LSTM language model (shallow fusion). Note that we consider this language model an external component to the ESPRESSO LSTM; we do not quantize it due to the lack of resources. However, we already show in our language modeling experiments that quantized language models retain their performance. We refer the reader to \citet{wang2019espresso} and training script\footnote{\url{https://github.com/freewym/espresso/blob/master/examples/asr_librispeech/run.sh}} for a complete description of the experimental setup.
  We initialize the quantized model from the pre-trained full-precision ESPRESSO LSTM. We train the quantized model for only 4 epochs due to the lack of resources. The quantized model is trained on 6 V100 GPUs where each epoch takes 2 days, so a total of 48 GPU days. The batch size is set to 8 mini-batch per GPU with maximum 8600 tokens. We made these changes because otherwise, the GPU would run out of VRAM due to the added fake quantization operations. For the first half of the first epoch, we gather statistics for quantization parameters then we enable QAT. The activation functions are swapped with quantization-aware PWL in the last epoch. The number of pieces for the quantization-aware PWLs is 96, except for the exponential function in the attention, which is 160 as we found out it was necessary to have more pieces because of its curvature. The number of pieces used is higher than in the language modeling experiments we did. However, the difference is that the inputs to the activation functions are 16-bit rather than 8-bit although the outputs are still quantized to 8-bit. It means we need more pieces to capture the inputs resolution better. Note that it would not be feasible to use a 16-bit Look-Up Table to compute the activation functions due to the size and cache misses, whereas using 96 pieces allows for a 170x reduction in memory consumption compared to LUT. 
\begin{table}
\begin{tabular}{ c | c | c }
      \hline 
	  Full-precision model & val & test\\
	  \hline
	  LayerNorm LSTM & $98.58 \pm 0.35$ & $ 94.84 \pm 0.21$ \\
      MadNorm LSTM & $\mathbf{97.20} \pm 0.47$ & $\mathbf{93.63} \pm 0.74$ \\ 
\end{tabular} 
\caption{Word-level perplexities on PTB for a full-precision LSTM with LayerNorm and a full-precision model with MadNorm.  Best results are averaged across 3 runs, and standard deviations are reported.}
\label{table:ptb-fp-ln-madnorm}
\end{table}

\begin{figure*}
\includegraphics[scale=0.7]{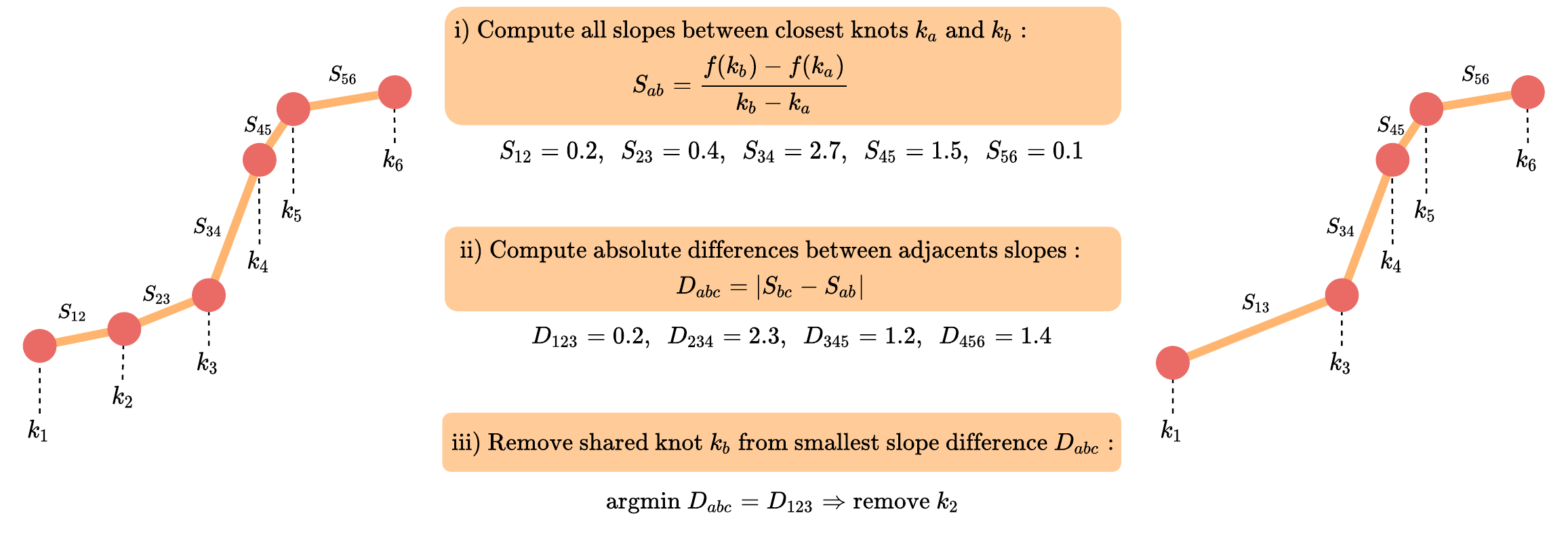}
\caption{Example of an iteration from our proposed quantization-aware PWL Algorithm \ref{alg:pwl}. The algorithm proceeds to reduce the number of pieces by merging two similar adjacents pieces. In this figure, the slopes $S_{12}$ and $S_{23}$ are the most similar pieces; therefore, the knot $k_2$ is removed.}
\label{fig:pwl-algo}
\end{figure*}


\begin{algorithm*}
{
\DontPrintSemicolon
\SetStartEndCondition{ }{}{}%
\SetKwProg{Fn}{def}{\string:}{}
\SetKwFunction{Range}{range}
\SetKw{KwTo}{in}\SetKwFor{For}{for}{\string:}{}%
\SetKwIF{If}{ElseIf}{Else}{if}{:}{elif}{else:}{}%
\SetKwFor{While}{while}{:}{fintq}%
\newcommand{\forcond}{$i=0$ \KwTo $n$}
\renewcommand{\forcond}{$i$ \KwTo\Range{$n$}}
\AlgoDontDisplayBlockMarkers\SetAlgoNoEnd\SetAlgoNoLine%
\SetKwFunction{SelectKnots}{select\_knots}%
\SetKwFunction{CreatePWL}{create\_quantization\_aware\_pwl}
\SetKwFunction{Len}{len}
\Fn{\SelectKnots{knots, intercepts, pwl\_nb}}{
dknots $\leftarrow$ knots$[1$:$]$ $-$ knots$[$:$-1]$ \\
dintercepts $\leftarrow$ intercepts$[1$:$]$ $-$ intercepts$[$:$-1]$\\
slopes $\leftarrow$ $\text{dintercepts}/\text{dknots}$\\
\If{\Len(slopes) $==$ pwl\_nb}{\Return{knots, slopes, intercepts}}
\Else{
diff\_adj\_slopes $\leftarrow \Big|$slopes$[$:$-1] -$ slopes$[1$:$]\Big|$\\
knot\_index\_to\_remove $\leftarrow \mathbf{argmin}$ diff\_adj\_slopes\\
remaining\_knots $\leftarrow$ knots.remove(knot\_index\_to\_remove)\\
remaining\_intercepts $\leftarrow$ intercepts.remove(knot\_index\_to\_remove)\\
\Return{\SelectKnots{$\text{remaining\_knots, remaining\_intercepts, pwl
\_nb}$
}
}
}
}
\;
\Fn{\CreatePWL{f, input\_scale, input\_zero\_point, b, pwl\_nb}}{
quantized\_knots $\leftarrow [0, ..., 2^b - 1]$ \tcp{Generate every $q_x$}
knots $\leftarrow \text{input\_scale} \odot (\text{quantized\_knots} - \text{input\_zero\_point})$ \tcp{Generate every $r_x$} 
intercepts $\leftarrow f(\text{knots})$\\
\{knots, slopes, intercepts\}  $\leftarrow$ \SelectKnots{$\text{knots, intercepts, pwl
\_nb}$}\\
\Return{knots, slopes, intercepts}
}
}
\caption{The algorithm recursively reduced the number of pieces until the wanted number of pieces is achieved. The algorithm needs to be provided the function to approximate $f$, the input scaling factor $S_x$ and zero-point $Z_x$, the quantization bitwidth $b$ and the number of linear pieces wanted. One iteration of select\_knots can be viewed in Figure \ref{fig:pwl-algo}.}
\label{alg:pwl}
\end{algorithm*}

\end{document}